\documentclass[10pt,twocolumn,letterpaper]{article}

\usepackage{cvpr}
\usepackage{times}
\usepackage{epsfig}
\usepackage{graphicx}
\usepackage{amsmath}
\usepackage{amssymb}

\usepackage{url}

\cvprfinalcopy 

\def\cvprPaperID{****} 

\begin{document}

\title{Image Retrieval with Fisher Vectors of Binary Features}

\author{Yusuke Uchida$^{\dagger, \ddagger}$ \\
$\dagger$The University of Tokyo\\
Tokyo, Japan\\
\and
Shigeyuki Sakazawa$^{\ddagger}$ \\
$\ddagger$KDDI R\&D Laboratories, Inc. \\
Saitama, Japan
\and
Shin'ichi Satoh$^{\dagger, \dagger\dagger}$ \\
$\dagger\dagger$National Institute of Informatics \\
Tokyo, Japan
}

\maketitle

\begin{abstract}
Recently,
the Fisher vector representation of local features
has attracted much attention because of its effectiveness
in both image classification and image retrieval.
Another trend in the area of image retrieval is
the use of binary features such as ORB, FREAK, and BRISK.
Considering the significant performance improvement for
accuracy in both image classification and retrieval by the Fisher vector of continuous feature descriptors,
if the Fisher vector were also to be applied to binary features,
we would receive similar benefits in binary feature based image retrieval and classification.
In this paper,
we derive the closed-form approximation of the Fisher vector of binary features modeled by the Bernoulli mixture model.
We also propose accelerating the Fisher vector by using the approximate value of posterior probability.
Experiments show that
the Fisher vector representation significantly improves the accuracy of image retrieval
compared with a bag of binary words approach.
\end{abstract}


\section{Introduction}
With the advancement of both
stable interest region detectors~\cite{mik_ijcv05} and
robust and distinctive descriptors~\cite{mik_pami05},
local feature based image or object retrieval has attracted a great deal of attention.
In local feature based image retrieval or recognition,
each image is first represented by a set of local features $X = \{ x_1, \cdots, x_t, \cdots, x_T \}$,
where $T$ is the number of local features.
The set of features $X$ is then encoded into a fixed length vector
in order to calculate any (dis)similarity between sets of features.
The most frequently used method is a bag-of-visual words (BoVW) representation~\cite{siv03},
where feature vectors are quantized into visual words (VWs) using a visual codebook that result in a histogram representation of VWs.

Recently,
the Fisher vector representation~\cite{per_cvpr07}
has attracted much attention because of its effectiveness.
The Fisher vector is defined by the gradient of log-likelihood function normalized with the Fisher information matrix.
In~\cite{per_cvpr07},
feature vectors are modeled by the Gaussian mixture model (GMM),
and a closed form approximation is first proposed for the Fisher information matrix of GMM.
Then,
the performance of the Fisher vector is improved in \cite{per_eccv10}
by using power and $\ell_2$ normalization.
Because the Fisher vector can represent higher order information than the BoVW representation,
it has been shown that it can outperform the BoVW representation
in both
image classification~\cite{per_eccv10}
and
image retrieval tasks~\cite{per_cvpr10, jeg_cvpr10, jeg_pami12}.

Another trend in the area of image retrieval is
the use of binary features such as
Oriented FAST and Rotated BRIEF (ORB)~\cite{rub_iccv11}, Fast Retina Keypoint (FREAK)~\cite{ala_cvpr12}, Binary Robust Invariant Scalable Keypoints (BRISK)~\cite{leu_iccv11},
KAZE features~\cite{alc_eccv12}, Accelerated-KAZE (A-KAZE)~\cite{alc_bmvc13}, Local Difference Binary (LDB)~\cite{yan_pami14}, and Learned Arrangements of Three patCH codes (LATCH)~\cite{lev_wacv16}.
Binary features are
one or two orders of magnitude faster than the Scale Invariant Feature Transform (SIFT)~\cite{low04} or Speeded Up Robust Features (SURF)~\cite{bay_cviu08} features in detection and description,
while providing comparable performance~\cite{rub_iccv11,hei_eccv12}.
These binary features are especially suitable for mobile visual search
or augmented reality on mobile devices\cite{yan_ismar12}.
While the Fisher vector is widely applied to continuous features (e.g., SIFT) that can be modeled by GMM,
to the best of our knowledge,
there has been no attempt to apply the Fisher vector to the abovementioned recent binary features for the purpose of image retrieval.
Considering the significant performance improvement for
accuracy in both image classification and retrieval by the Fisher vector of continuous features,
if the Fisher vector were also to be applied to binary features,
we would receive similar benefits in binary feature-based image retrieval and classification.

In this paper,
we propose to apply the Fisher vector representation to binary features to improve the accuracy of binary feature based image retrieval.
Table~\ref{tab:position} shows the position of this paper.
Our main contribution is to model binary features using the Bernoulli mixture model (BMM)
and derive the closed-form approximation of the Fisher vector of BMM~\cite{uch_acpr13}.
Experimental results show that the proposed Fisher vector outperforms the BoVW method on various types of objects.
In addition, we also propose a fast approximation method to accelerate the computation of the proposed Fisher vectors by one order of magnitude with comparable performance.
In the experiments,
we evaluate the effectiveness of both the proposed Fisher vector representation of binary features and their associated vector normalization method.
In particular,
we demonstrate that a normalization method, originally proposed for the other vector representation, also works well for the proposed Fisher vector.
The proposed Fisher vector representation of binary features is general and not restricted to image features;
it is also expected to be applicable to other modalities
such as audio signals~\cite{hai02, ang_icme12}.

The rest of this paper is organized as follows.
In Section 2,
the recent binary features that we are going to model are briefly introduced.
In Section 3,
we describe the BoVW and Fisher vector image representations,
which have been applied to continuous feature vectors (e.g., SIFT).
In Section 4,
we model binary features with BMM
and derive the Fisher vector of BMM,
which enables us to apply the Fisher vector representation to binary features.
In Section 5,
the effectiveness of the Fisher vector of binary features is confirmed.
Our conclusions are presented in Section 6.

\begin{table}[tb]
	\centering
	\caption{Position of this paper.}
	\label{tab:position}
	\begin{tabular}{c|cc} \hline
		Feature type		& BoVW			& Fisher Vector	\\ \hline
		Continuous	& \cite{siv03}	& \cite{per_cvpr07}	\\
		Binary		& \cite{gal_iros11}	& This paper	\\ \hline
	\end{tabular} \\
\end{table}

\section{Local binary features}
Recently,
binary features such as ORB~\cite{rub_iccv11}, FREAK~\cite{ala_cvpr12}, and BRISK~\cite{leu_iccv11}
have attracted much attention~\cite{hei_eccv12}.
Binary features are
one or two orders of magnitude faster than SIFT or SURF features in extraction,
while providing comparable performance to SIFT and SURF.
In this section,
recent binary features are briefly introduced.

\subsection{Detection}
Most of the local binary features employ fast feature detectors.
The ORB feature utilizes the Features from Accelerated Segment Test (FAST)~\cite{ros_iccv05} detector,
which detects pixels that are brighter or darker than neighboring pixels based on the accelerated segment test.
The test is optimized to reject candidate pixels very quickly,
realizing extremely fast feature detection.
In order to ensure approximate scale invariance,
feature points are detected from an image pyramid.
The FREAK and BRISK features
adopt the multi-scale version of the Adaptive and Generic Accelerated Segment Test (AGAST)~\cite{mai_eccv10} detector.
Although the AGAST detector is based on the same criteria as FAST,
the detection is accelerated by using an optimal decision tree in deciding
whether each pixel satisfies the criteria or not.

\subsection{Description}
\label{sec:desc}
Local binary features extract binary strings from patches of interest regions
instead of extracting gradient-based high-dimensional feature vectors like SIFT.
Many methods utilize binary tests in extracting binary strings.
The BRIEF descriptor~\cite{cal_eccv10}, a pioneering work in the area of binary descriptors,
is a bit string description of an image patch constructed from a set of binary intensity tests.
Consider the $t$-th smoothed image patch $p_t$,
a binary test $\tau$ for $d$-th bit is defined by:
\begin{equation}
	x_{td} = \tau(p_t; a_d, b_d) =
	\begin{cases}
		\, 1 & \mathrm{if} \; p_t(a_d) \ge p_t(b_d) \\
		\, 0 & \mathrm{else}
	\end{cases},
\end{equation}
where
$a_d$ and $b_d$ denote relative positions in the patch $p_t$,
and $p_t(\cdot)$ denotes the intensity at the point.
Using $D$ independent tests,
we obtain $D$-bit binary string $x_t = (x_{t1}, \cdots, x_{td}, \cdots, x_{tD})$ for the patch $p_t$.
The ORB feature employs
a learning method for de-correlating BRIEF features
under rotational invariance.
Although the BRISK and FREAK features use
different sampling patterns from BRIEF,
they are also based on a set of binary intensity tests.
These binary features are designed so that
each bit has the same probability of being 1 or 0,
and bits are uncorrelated.

In addition to these methods that extract binary features directly,
there are many methods that encode
continuous feature vectors (e.g., SIFT) into compact binary codes~\cite{rag_nips09, wan10, gon_cvpr11, amb_iccv11, lee_accv12, iri_cvpr14, lio_cvpr15}.
By using these methods,
an image can be represented as a set of binary features.

\section{Image representations}
In local feature based image retrieval or recognition,
each image is first represented by a set of local features $X = \{ x_1, \cdots, x_t, \cdots, x_T \}$.
A set of features $X$ is then encoded into a fixed length vector
in order to calculate (dis)similarity between sets of features.
In this section,
two encoding methods are introduced.

\subsection{Bag-of-Visual Words}
The BoVW framework is the de-facto standard to encode local features into a fixed length vector.
In the BoVW framework,
feature vectors are quantized into VWs using a visual codebook that result in a histogram representation of VWs.
Image (dis)similarity is measured by $L_1$ or $L_2$ distance between the normalized histograms.
Although it was first proposed for an image retrieval task~\cite{siv03},
it is now widely used for both
image retrieval~\cite{nis06, phi07, jeg_ijcv10, uch_icpr12}
and
image classification~\cite{laz_cvpr06, jia_civr07}.
In~\cite{gal_iros11},
the bag-of-visual words approach is also applied to binary features.

\subsection{Fisher Kernel and Fisher Vector}
The Fisher kernel is a powerful tool for combining the benefits of generative and discriminative approaches~\cite{jaa_nips98}.
Let $X = \{ x_1, \cdots, x_t, \cdots, x_T \}$ denote
the set of $T$ local feature vectors extracted from an image.
We assume that the generation process of $X$
can be modeled by a probability density function $p(X | \lambda)$ whose parameters are denoted by $\lambda$.
In~\cite{jaa_nips98},
it is proposed to describe $X$ by
the gradient $G^{X}_{\lambda}$ of the log-likelihood function,
which is also referred to as the Fisher score:
\begin{equation}
\label{eq:1}
G^{X}_{\lambda} = \frac{1}{T}
\nabla_{\lambda} \mathcal{L} (X | \lambda),
\end{equation}
where $\mathcal{L} (X | \lambda)$ denotes the log-likelihood function:
\begin{equation}
\mathcal{L} (X | \lambda) = \log p(X | \lambda).
\end{equation}
The gradient vector describes the direction in which parameters should be modified to best fit the data~\cite{per_cvpr07}.
A natural kernel on these gradients is the Fisher kernel~\cite{jaa_nips98},
which is based on the idea of natural gradient~\cite{ama_nc98}:
\begin{equation}
K(X, Y) = G^{X}_{\lambda} F_{\lambda}^{-1} G^{Y}_{\lambda}.
\end{equation}
$F_{\lambda}$ is the Fisher information matrix of $p(X | \lambda)$ defined as
\begin{equation}
\label{eq:fim}
F_{\lambda} =
\mathrm{E}_X [ \nabla_{\lambda} \mathcal{L} (X | \lambda) \; \nabla_{\lambda} \mathcal{L} (X | \lambda)^{\mathrm{T}}].
\end{equation}
Because $F_{\lambda}^{-1}$ is positive semidefinite and symmetric,
it has a Cholesky decomposition $F_{\lambda}^{-1} = L_{\lambda}^{\mathrm{T}} L_{\lambda}$.
Therefore the Fisher kernel is rewritten as a dot-product between normalized gradient vectors $\mathcal{G}_{\lambda}^X$ with:
\begin{equation}
\label{eq:fv}
\mathcal{G}_{\lambda}^X = L_{\lambda} G^{X}_{\lambda}.
\end{equation}
The normalized gradient vector $\mathcal{G}_{\lambda}^X$ is referred to as the Fisher vector of $X$~\cite{per_eccv10}.

In~\cite{per_cvpr07},
the generation process of feature vectors (SIFT) are modeled by GMM,
and
the diagonal closed-form approximation of the Fisher vector is derived.
Then,
the performance of the Fisher vector is significantly improved in \cite{per_eccv10}
by using power normalization and $\ell_2$ normalization.
The Fisher vector framework has achieved promising results and is becoming the new standard
in both
image classification~\cite{per_eccv10}
and
image retrieval tasks~\cite{per_cvpr10, jeg_cvpr10, jeg_pami12}.
There are several extensions to this framework
such as multiple-layered Fisher vector~\cite{sim_nips13} and
a combination with Convolutional Neural Networks (CNN)~\cite{per_cvpr15}.

While the Fisher vector is widely applied to continuous features (e.g., SIFT) that can be modeled by GMM,
to the best of our knowledge,
there has been no attempt to apply the Fisher vector to recent binary features such as ORB~\cite{rub_iccv11} for the purpose of image retrieval.
In this paper,
we derive the closed-form approximation of the Fisher vector of binary features which are modeled by the Bernoulli mixture model,
and evaluate the effectiveness of both the Fisher vector of binary features and their associated normalization approaches.

\subsection{Vector of Locally Aggregated Descriptors}
\label{sec:vlad}
In \cite{jeg_cvpr10},
J{\'e}gou et al. proposed an efficient way of aggregating local features into a vector of fixed dimension,
namely Vector of Locally Aggregated Descriptors (VLAD).
In the construction of VLAD,
each descriptor is first assigned to the closest visual word in a visual codebook in the same way
as in the construction of the BoVW vector.
For each of the visual words,
the residuals in quantization are accumulated,
and the sums of residuals are concatenated into a single vector, VLAD.
VLAD can be considered as the simplified non-probabilistic version of the Fisher vector.
VLAD has been further improved by modifying its vector normalization or aggregation step~\cite{ara_cvpr13, spy_tmm14}.
As the performance of VLAD is about the same or a little worse than the Fisher vector~\cite{jeg_pami12},
we focus on the Fisher vector in this paper.

\section{Fisher Vector for Binary Features}
In this section,
we model binary features with the Bernoulli distribution,
and derive the Fisher vector representation of binary features.

\subsection{Bernoulli Mixture Model}
Let $x_t$ denote a $D$-dimensional binary feature out of $T$ binary features $X = \{ x_1, \cdots, x_t, \cdots, x_T \}$ extracted from an image.
In modeling binary features,
it is straightforward to adopt a single multivariate Bernoulli distribution.
However,
although many binary descriptors are designed
so that bits of resulting binary features are uncorrelated~\cite{rub_iccv11},
there are still strong dependencies among the bits.
Therefore, a single multivariate Bernoulli component will
be inadequate to cope with the kind of complex bit dependencies that often underlie binary features.
This drawback is overcome when several Bernoulli components are adequately mixed.
In this paper,
we model binary features with the Bernoulli mixture model (BMM).
The use of BMM instead of a single multivariate Bernoulli distribution
will be justified in the experimental section.

Let $\lambda = \{ w_i, \mu_{id}, i = 1, \cdots, N, d = 1, \cdots, D \}$ denote a set of parameters for a multivariate Bernoulli mixture model with $N$ components,
and $x_{td}$ represents the $d$-th bit of $x_t$.
Given the parameter set $\lambda$,
the probability density function of $T$ binary features $X$ is described as:
\begin{eqnarray}
p(X | \lambda) &=& \prod_{t =1}^T p(x_t | \lambda), \nonumber \\
p(x_t | \lambda) &=& \sum_{i =1}^N w_i p_i (x_t | \lambda), \nonumber \\
p_i (x_t | \lambda) &=& \prod_{d = 1}^D \mu_{id}^{x_{td}} (1 - \mu_{id})^{1 - x_{td}}.
\end{eqnarray}

In order to estimate the values of the parameter set $\lambda$,
given a set of training binary features $x_1, \cdots, x_s, \cdots, x_S$,
the expectation-maximization (EM) algorithm is applied~\cite{jua_icpr04}.
In the expectation step,
the occupancy probability $\gamma_s (i)$ (or posterior probability $p(i | x_s, \lambda)$)
of $x_s$ being generated by the $i$-th component of BMM is calculated as
\begin{equation}
\label{eq:gamma}
\gamma_s (i) = p(i | x_s, \lambda) = \frac{w_i p_i (x_s | \lambda)}{\sum_{j = 1}^N w_j p_j (x_s | \lambda)}.
\end{equation}
In the maximization step, the parameters are updated as
\begin{equation}
S_i = \sum_{s=1}^S \gamma_s (i), \;
w_i = S_i / S, \;
\mu_{id} =
\frac{1}{S_i} \sum_{s=1}^S \gamma_s (i) x_{sd}.
\end{equation}
In our implementation,
parameter $w_i$ is initialized with $1 / N$,
and $\mu_{id}$ is with uniform distribution $U(0.25, 0.75)$.
From our experience,
these initial parameters do not have a large impact on the final result.

\subsection{Deriving the Fisher Vector of BMM}
In this section,
we derive the Fisher vector of BMM.
In order to calculate the Fisher vector $\mathcal{G}_{\lambda}^X$ in Eq.~(\ref{eq:fv}),
the Fisher score $G_{\lambda}^X$ in Eq.~(\ref{eq:1})
and
the Fisher information matrix $F_{\lambda}$ in Eq.~(\ref{eq:fim})
should be calculated.

Letting $G_{\mu_{id}}^X$ denote the Fisher score w.r.t. the parameter $\mu_{id} \in \lambda$,
$G_{\mu_{id}}^X$ is calculated as:
\begin{eqnarray}
G_{\mu_{id}}^X &=&
\frac{1}{T} \frac{\partial \mathcal{L} (X | \lambda)}{\partial \mu_{id}} =
\frac{1}{T} \sum_{t = 1}^T \frac{\partial \mathcal{L} (x_t | \lambda)}{\partial \mu_{id}} \nonumber \\
&=&
\frac{1}{T} \sum_{t = 1}^T
\frac{1}{p_i (x_t | \lambda)}
\frac{\partial p_i (x_t | \lambda)}{\partial \mu_{id}},
\end{eqnarray}
where
\begin{equation}
\frac{\partial p_i (x_t | \lambda)}{\partial \mu_{id}} =
(- 1)^{1 - x_{td}} \prod_{e = 1, e \neq d}^D \mu_{ie}^{x_{te}} (1 - \mu_{ie})^{1 - x_{te}}.
\end{equation}
Finally we obtain:
\begin{equation}
\label{eq:fs}
G_{\mu_{id}}^X =
\frac{1}{T} \sum_{t = 1}^T
\gamma_t (i) \frac{(-1)^{1 - x_{td}}}{\mu_{id}^{x_{td}} (1 - \mu_{id})^{1 - x_{td}}},
\end{equation}
where $\gamma_t (i)$ is the occupancy probability defined in Eq.~(\ref{eq:gamma}).

Then,
we derive the approximate Fisher information matrix of BMM
under the following three assumptions~\cite{per_cvpr07}:
(1) the Fisher information matrix $F_{\lambda}$ is diagonal,
(2) the number of binary features $x_t$ extracted from an image is constant and equal to $T$, and
(3) the occupancy probability $\gamma_t (i)$ is peaky;
there is one index $i$ such that $\gamma_t (i) \approx 1$ and that $\forall j \neq i$, $\gamma_t (j) \approx 0$.

As we assume the Fisher information matrix is diagonal,
Eq.~(\ref{eq:fim}) is approximated as
$F_{\lambda} \approx \mathrm{diag}(F_{\mu_{11}}, \cdots, F_{\mu_{ND}})$,
where
$F_{\mu_{id}}$ denotes the Fisher information w.r.t. $\mu_{id}$:
\begin{equation}
\label{eq:fi}
F_{\mu_{id}}
=
\mathrm{E} \left[ \left(
\frac{\partial \mathcal{L} (X | \lambda)}{\partial \mu_{id}}
\right)^2 \right].
\end{equation}
Then, with the (2) and (3) assumptions,
we approximately obtain:
\begin{equation}
\label{eq:fi2}
F_{\mu_{id}} =
T w_i
\left(
\frac{\sum_{j=1}^N w_j \mu_{jd}}{\mu_{id}^2}
+
\frac{\sum_{j=1}^N w_j (1 - \mu_{jd})}{(1 - \mu_{id})^2}
\right).
\end{equation}
Please refer to Appendix for the derivation. 
Finally,
the Fisher vector $\mathcal{G}_{\lambda}^X$ is obtained with the concatenation of normalized Fisher scores $F_{\mu_{id}}^{-1/2} G_{\mu_{id}}^X \; (i = 1, \cdots, N, d = 1, \cdots, D)$.

\subsection{Vector Normalization}
\label{sec:normalization}
The Fisher vector is further normalized with
power normalization and $\ell_2$ normalization~\cite{per_eccv10}.
Given a Fisher vector $z = \mathcal{G}_{\lambda}^X$,
the power-normalized vector $f(z)$ is calculated as
\begin{equation}
	f(z) = \mathrm{sign}(z)|z|^{\alpha}.
\end{equation}
In experiments,
we set $\alpha = 0.5$ as recommended in~\cite{per_eccv10}.
After the power normalization,
$\ell_2$ normalization is performed to $f(z)$,
resulting in the final Fisher vector representation of the set of binary features.
In addition,
we propose to use intra normalization~\cite{ara_cvpr13} for this Fisher vector
instead of the power and $\ell_2$ normalization.
The intra normalization method was originally proposed for the VLAD representation described in Section~\ref{sec:vlad}, not for the Fisher vector.
However,
the purpose of intra normalization is to alleviate the problem of \textit{burstiness} in visual words~\cite{jeg_cvpr09, ara_cvpr13}
and, this is the same as that of power normalization.
Therefore,
it is also expected to work well for the Fisher vector.
In the case of the Fisher vector,
intra normalization is done
by performing $\ell_2$ normalization within each BMM component.

\subsection{Fast Approximated Fisher Vector}
\label{sec:acceleration}
The most computationally expensive part of the proposed Fisher vector is
the calculation of the occupancy probability $\gamma_t (i)$ in Eq.~(\ref{eq:fs})
because $F_{\mu_{id}}$ in Eq.~(\ref{eq:fi2}) does not depend on the input vector $X$
and can be precomputed.
In this paper,
we also propose to accelerate the proposed Fisher vector by using the approximate value of $\gamma_t (i)$.

Firstly,
each $i$-th component of BMM is converted into a representative binary vector $y_i = (y_{i1}, \cdots, y_{iD})$ as
\begin{equation}
\label{eq:approx1}
	y_{id} =
	\begin{cases}
		\, 1 & \mu_{id} \ge 0.5 \\
		\, 0 & \mu_{id} < 0.5.
	\end{cases}
\end{equation}
Then,
for each $x_t \in X$,
the most similar representative binary vector $y_{\hat{i}}$ is calculated by
$\hat{i} = \arg\min_i |x_t - y_i|$.
This involves only the calculation of Hamming distance and can be done very fast.
Finally, we obtain approximated $\gamma_{t}' (i)$ as
\begin{equation}
\label{eq:approx2}
	\gamma_{t}' (i) =
	\begin{cases}
		\, 1 & i = \hat{i} \\
		\, 0 & i \neq \hat{i}.
	\end{cases}
\end{equation}
This approximation is also based on the assumption that
the occupancy probability $\gamma_t (i)$ is peaky.

\begin{figure}[tb]
	\centering
	\begin{minipage}[b]{0.24\linewidth}
		\includegraphics[width=\linewidth]{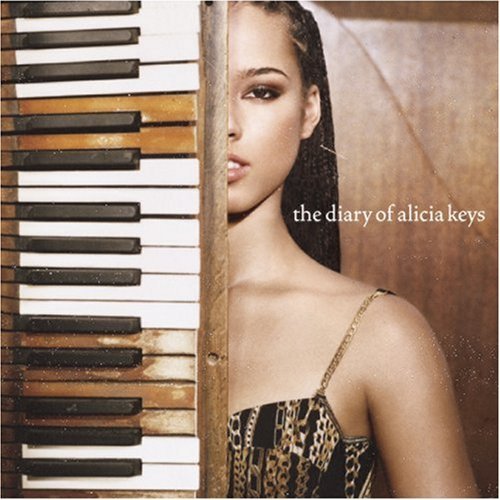}
	\end{minipage}
	\begin{minipage}[b]{0.24\linewidth}
		\includegraphics[width=\linewidth]{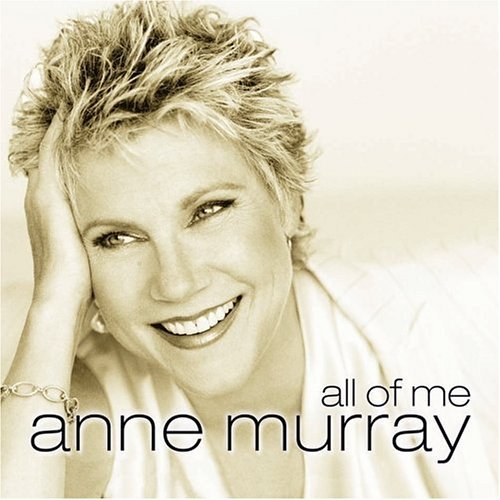}
	\end{minipage}
	\begin{minipage}[b]{0.24\linewidth}
		\includegraphics[width=\linewidth]{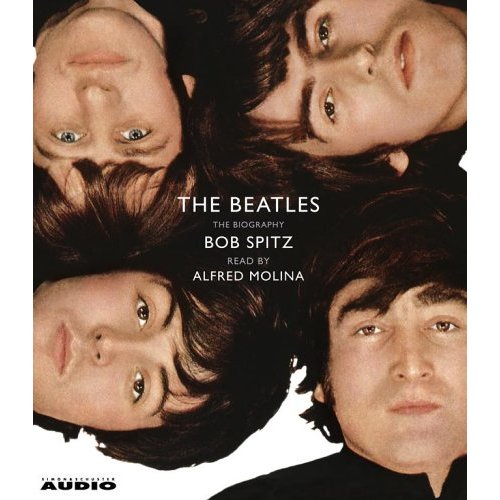}
	\end{minipage}
	\begin{minipage}[b]{0.24\linewidth}
		\includegraphics[width=\linewidth]{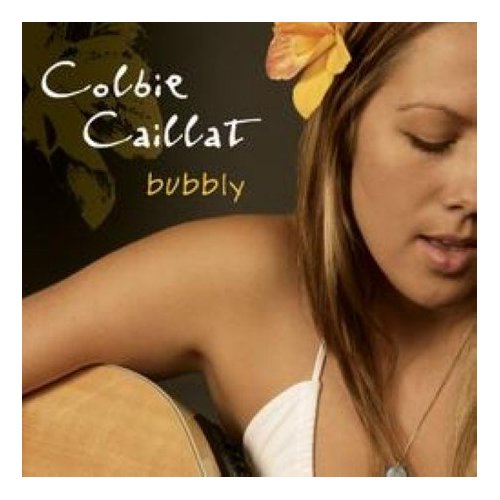}
	\end{minipage}
	\begin{minipage}[b]{0.24\linewidth}
		\includegraphics[width=\linewidth]{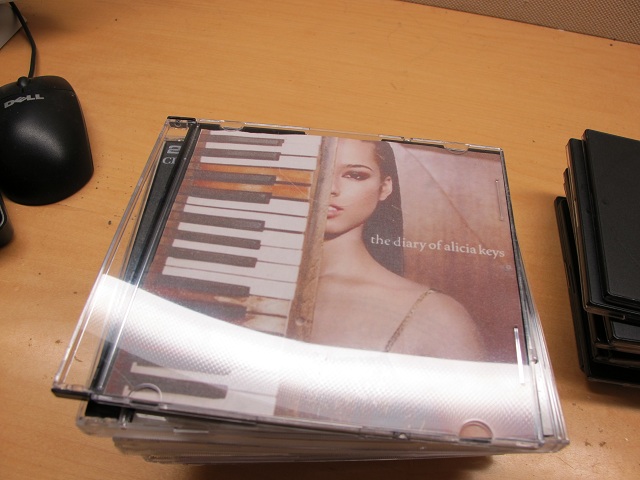}
	\end{minipage}
	\begin{minipage}[b]{0.24\linewidth}
		\includegraphics[width=\linewidth]{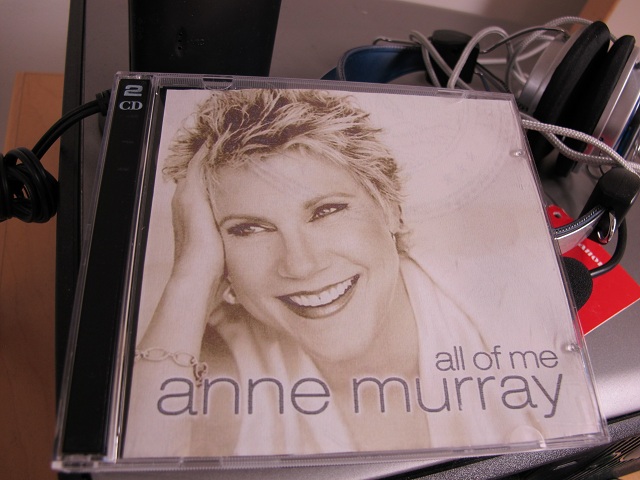}
	\end{minipage}
	\begin{minipage}[b]{0.24\linewidth}
		\includegraphics[width=\linewidth]{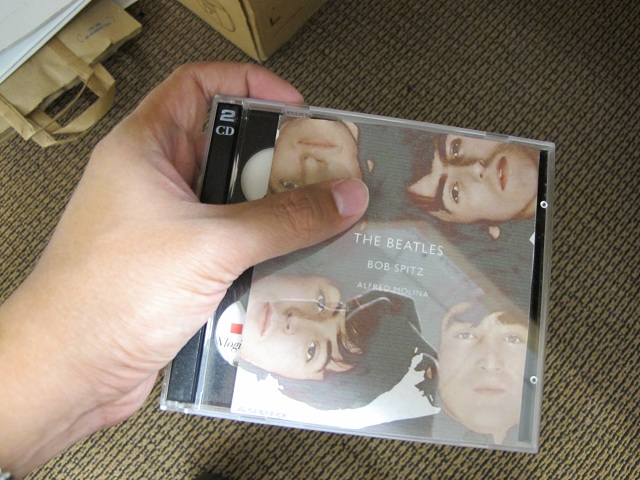}
	\end{minipage}
	\begin{minipage}[b]{0.24\linewidth}
		\includegraphics[width=\linewidth]{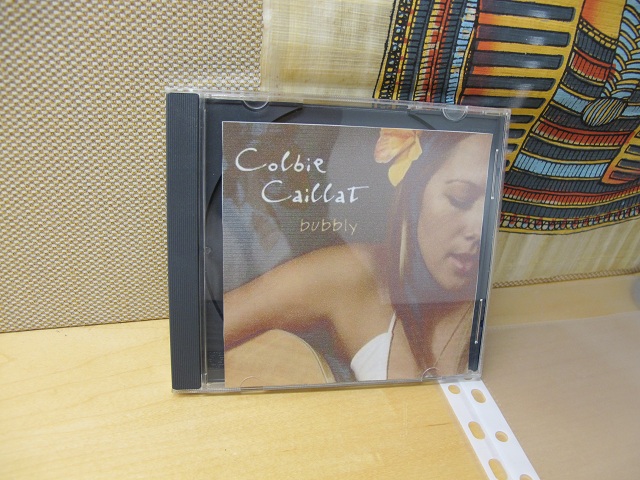}
	\end{minipage} \\
	\caption{Example images from the Stanford MVS dataset.
	Images in the top row are examples of reference images and images in the bottom row are examples of query images.}
	\label{fig:sample}
\end{figure}

\section{Experiment}
In the experiments,
the Stanford mobile visual search dataset\footnote{\url{http://www.stanford.edu/~dmchen/mvs.html}} is used
to evaluate the effectiveness of the proposed Fisher vector in image retrieval.
The dataset contains camera-phone images of CD covers, books, business cards, DVD covers, outdoor landmarks, museum paintings, print documents, and video clips.
These images consist of 100 reference images and 400 query images.
Because some query images are too large (10M pixels),
all images are resized so that
the longest sides of the images are less than 640 pixels, while keeping the original aspect ratio.
Figure~\ref{fig:sample} shows example images from the dataset.

The dissimilarity between two images is defined by the Euclidean distance between either the BoVW or the Fisher vector representations of the images.
As an indicator of the retrieval performance,
mean average precision (MAP)~\cite{nis06, jeg_ijcv10} is used.
For each query,
a precision-recall curve is obtained based on the retrieval results.
Average precision is calculated as the area under the precision-recall curve.
Finally, the MAP score is calculated as the mean of the average precisions over all queries.

As a binary feature,
we adopt the ORB~\cite{rub_iccv11} descriptor,
which is one of the most frequently used binary features.
An implementation of the ORB descriptor is available in an open source library\footnote{\url{http://opencv.org/}}.
On average, 900 features are extracted from four scales.
The parameter set $\lambda$ is estimated with the EM algorithm
using one million ORB binary features extracted from
the MIR Flickr collection\footnote{\url{http://press.liacs.nl/mirflickr/}}.
The following experiments were performed on a standard desktop PC with a Core i7 970 CPU.

\begin{figure*}[tb]
	\centering
	\begin{minipage}[c]{0.19\linewidth}
		\includegraphics[width=\linewidth]{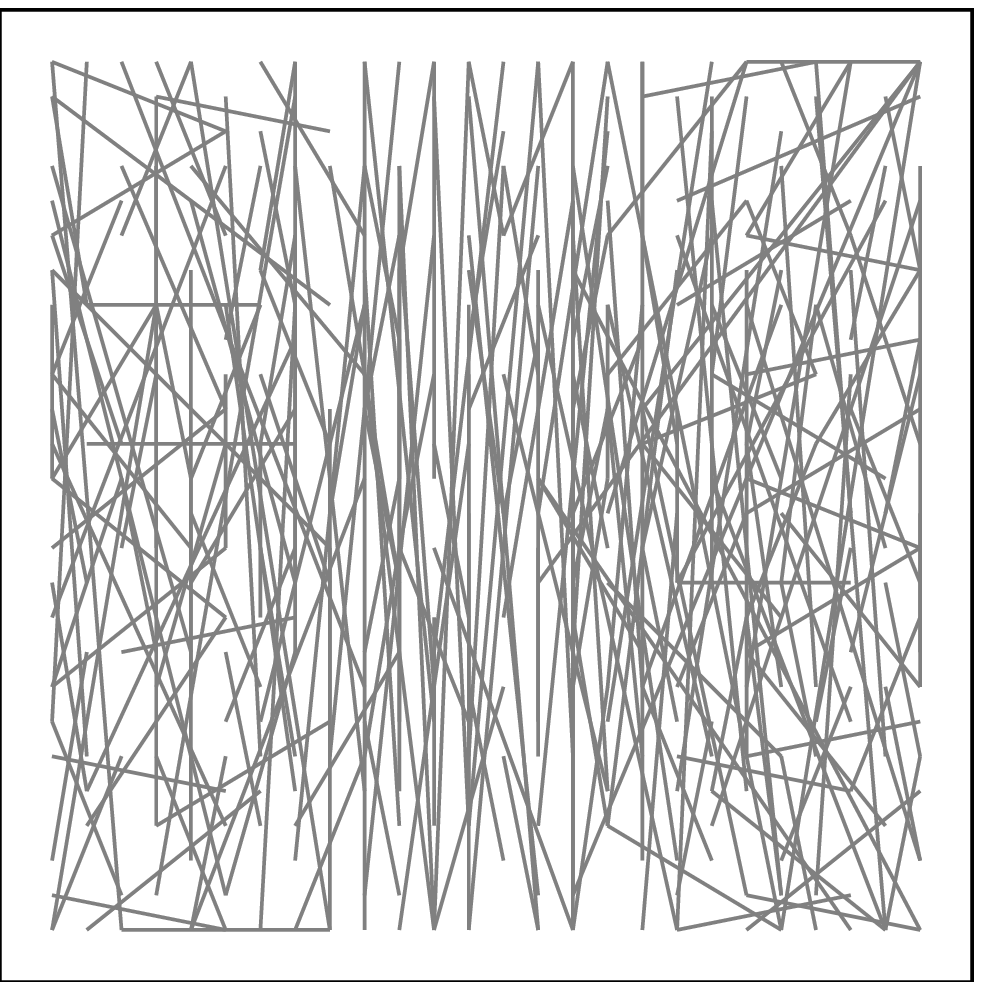} \\
		\centering (a) \\
	\end{minipage}
	\begin{minipage}[c]{0.19\linewidth}
		\includegraphics[width=\linewidth]{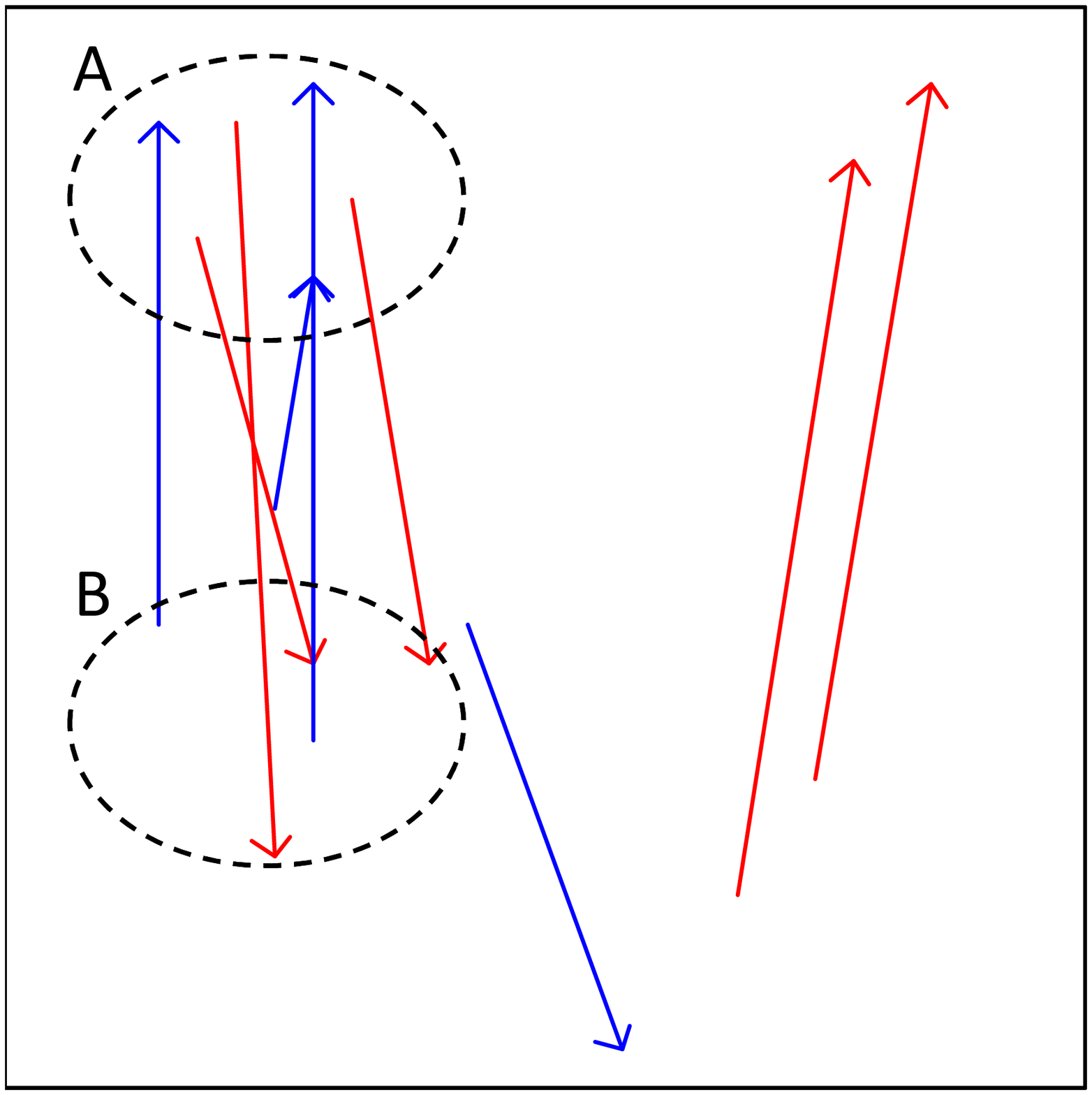} \\
		\centering (b) \\
	\end{minipage}
	\begin{minipage}[c]{0.19\linewidth}
		\includegraphics[width=\linewidth]{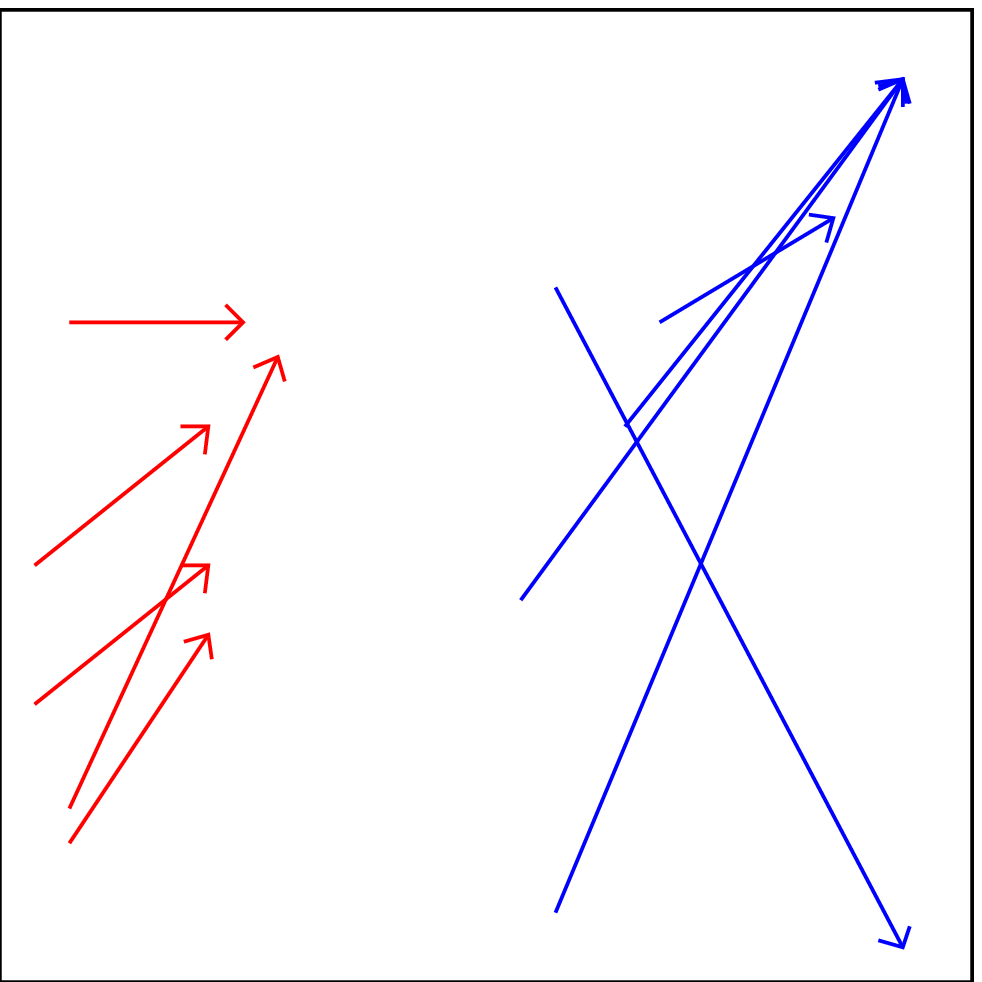} \\
		\centering (c) \\
	\end{minipage}
	\begin{minipage}[c]{0.19\linewidth}
		\includegraphics[width=\linewidth]{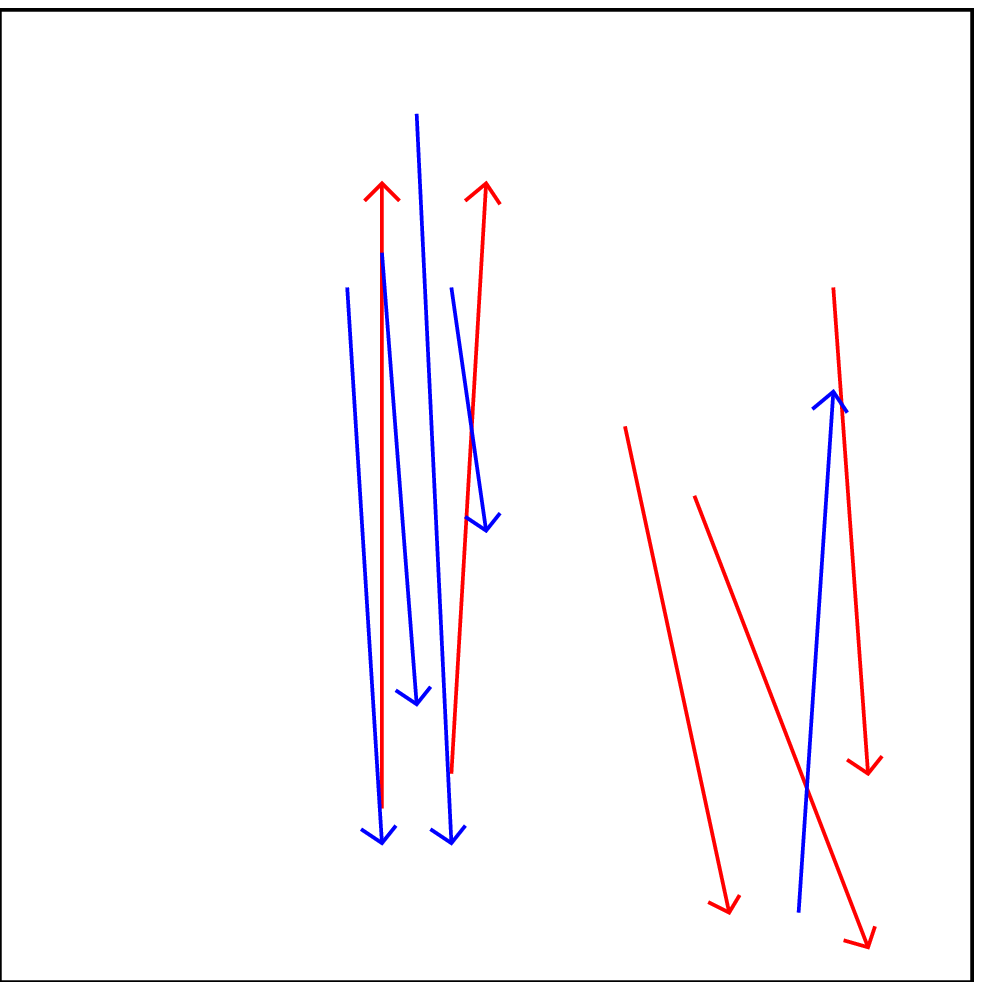} \\
		\centering (d) \\
	\end{minipage}
	\begin{minipage}[c]{0.19\linewidth}
		\includegraphics[width=\linewidth]{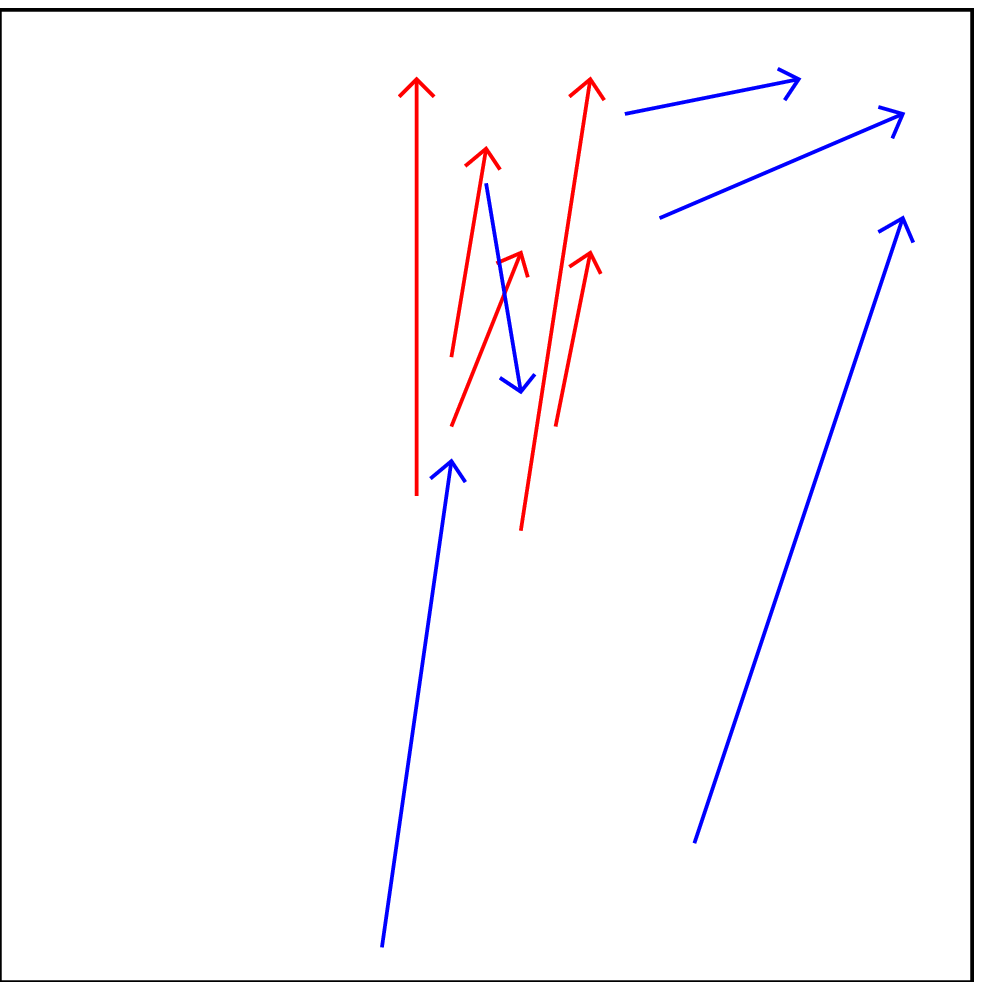} \\
		\centering (e) \\
	\end{minipage}
	\centering
	\caption{(a) All point pairs of 256 binary tests used in the ORB descriptor.
	(b)-(e) Five tests corresponding to the bits with the top five probabilities $\mu_{id}$ of being 1 (red) and 0 (blue).
	Four randomly selected components out of $N=32$ components are shown.}
	\label{fig:pairs}
\end{figure*}

\subsection{Clustering Effect}
First,
we investigate
the clustering results generated from the estimation of the parameter set $\lambda$ of BMM with $N = 32$ components.
Figure~\ref{fig:pairs} (a) represents
all point pairs $(a_d, b_d)$ of the 256 binary tests used in the ORB descriptor
explained in Section~\ref{sec:desc}.
Figures~\ref{fig:pairs} (b)-(e) visualize a part of the parameter sets $\lambda$ of
four randomly selected components out of $N=32$ components.
In each figure, red (blue) arrows represent five tests corresponding to
the five largest (smallest) $\mu_{id}$.
The arrows are drawn from $a_d$ to $b_d$, and
$\mu_{id}$ represents the probability that $a_d$ is brighter than $b_d$.
Therefore, the pixel at the head of a red arrow tends to be brighter than the tail of the red arrow
while the pixel at the head of a blue arrow tends to be darker than the tail of the blue arrow.
We can see that the binary tests with the largest and the smallest $\mu_{id}$ concentrate on small areas
(e.g., between the areas A and B in Figure~\ref{fig:pairs} (b)).
Thus, Figure~\ref{fig:pairs} implies that
some bits of the ORB descriptor are highly correlated
and that BMM successfully captures this correlation.
The result justifies
the use of BMM instead of the single multivariate Bernoulli distribution
to model binary features.

\begin{figure*}[tb]
	\centering
	\begin{minipage}[b]{0.39\linewidth}
		\includegraphics[width=\linewidth]{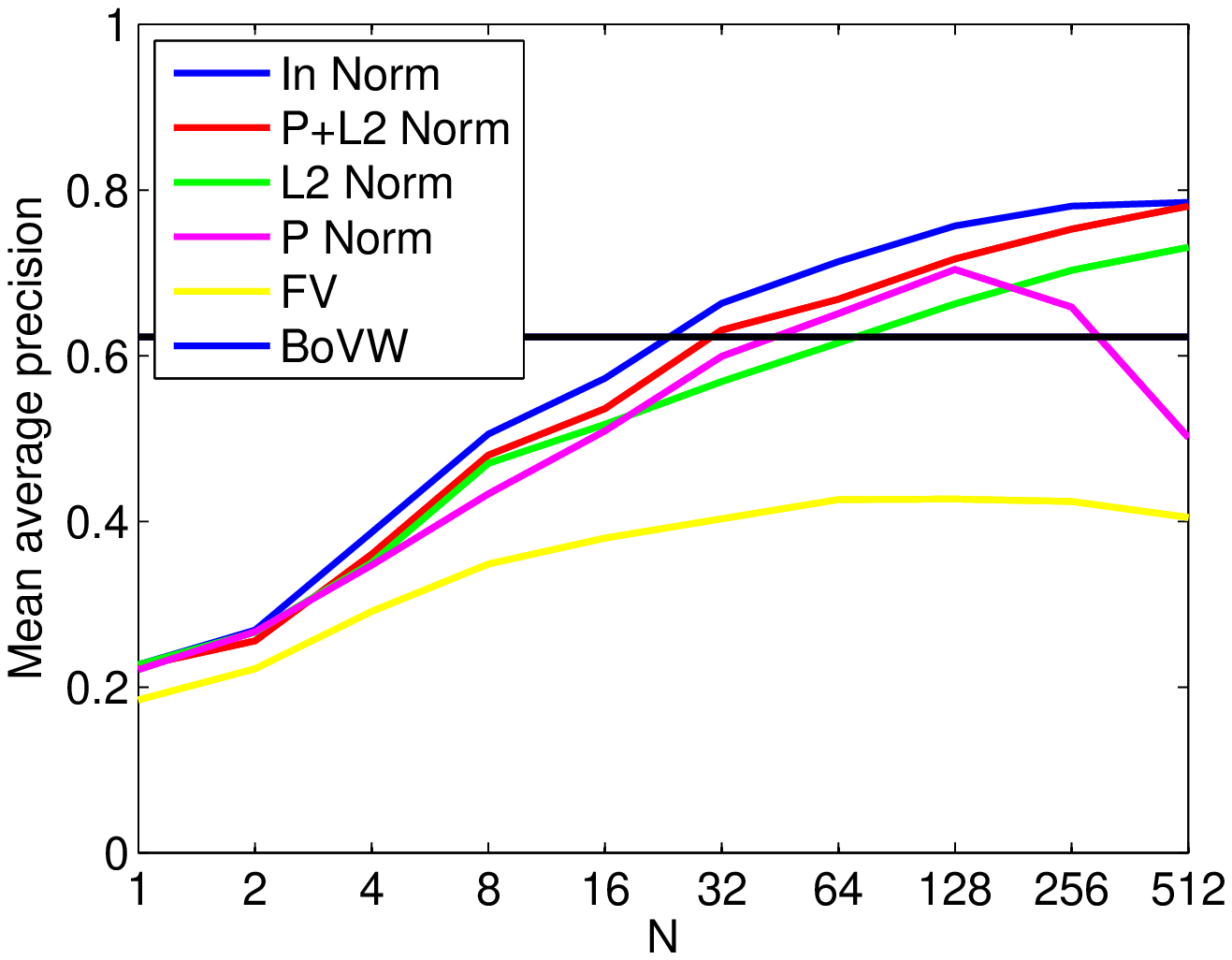} \\
		\centering (a) cd
	\end{minipage}
	\begin{minipage}[b]{0.39\linewidth}
		\includegraphics[width=\linewidth]{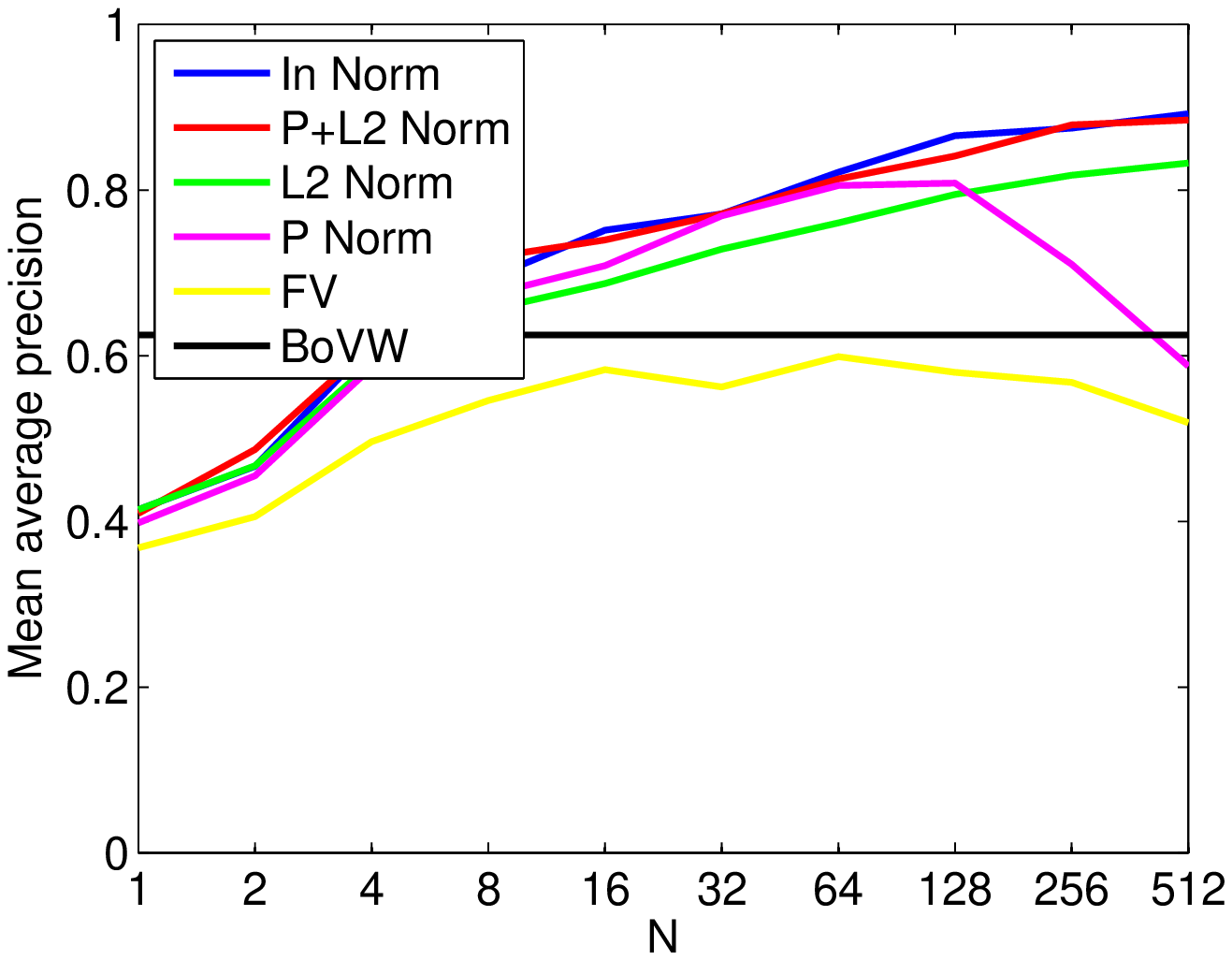} \\
		\centering (b) book
	\end{minipage}
	\begin{minipage}[b]{0.39\linewidth}
		\includegraphics[width=\linewidth]{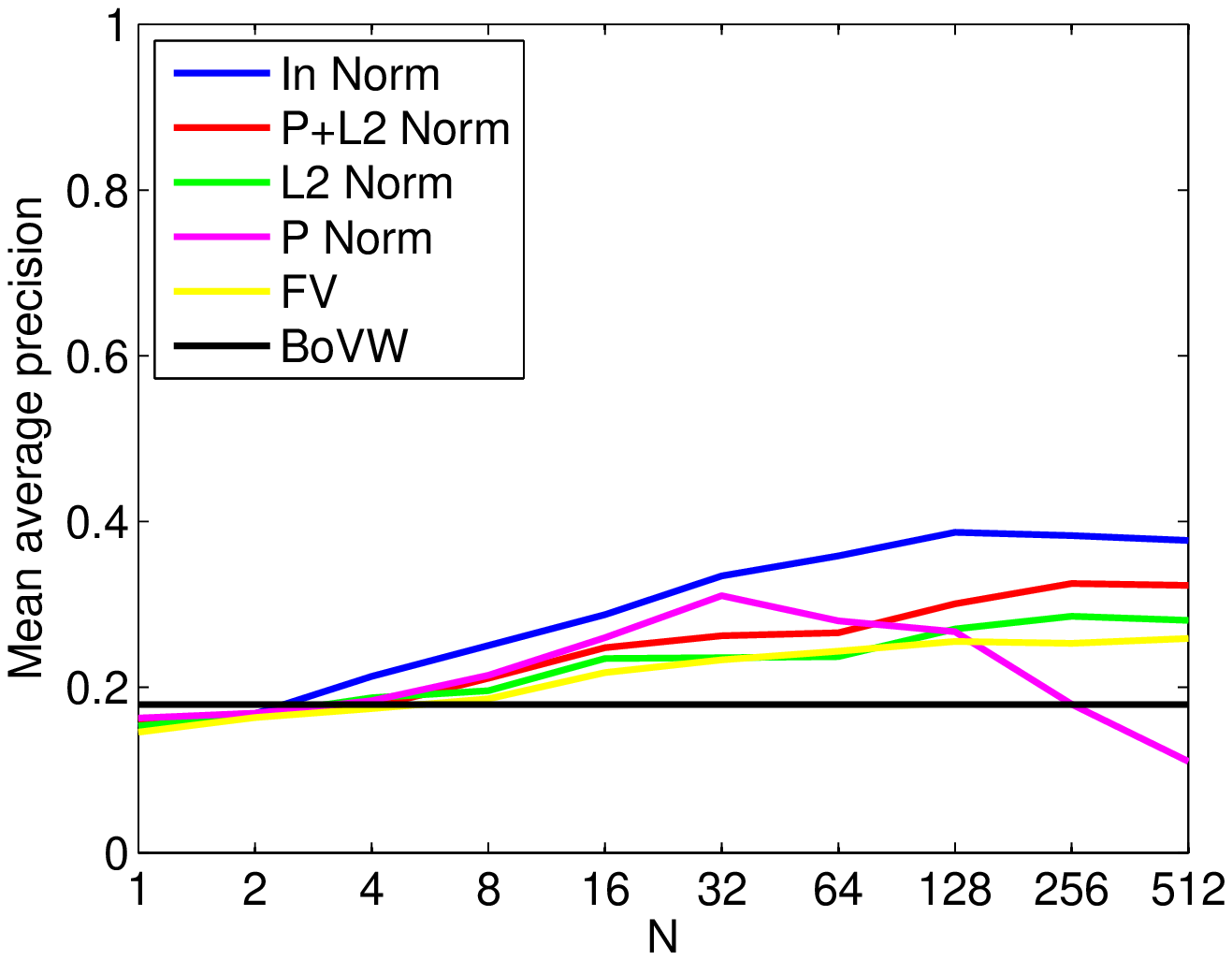} \\
		\centering (c) card
	\end{minipage}
	\begin{minipage}[b]{0.39\linewidth}
		\includegraphics[width=\linewidth]{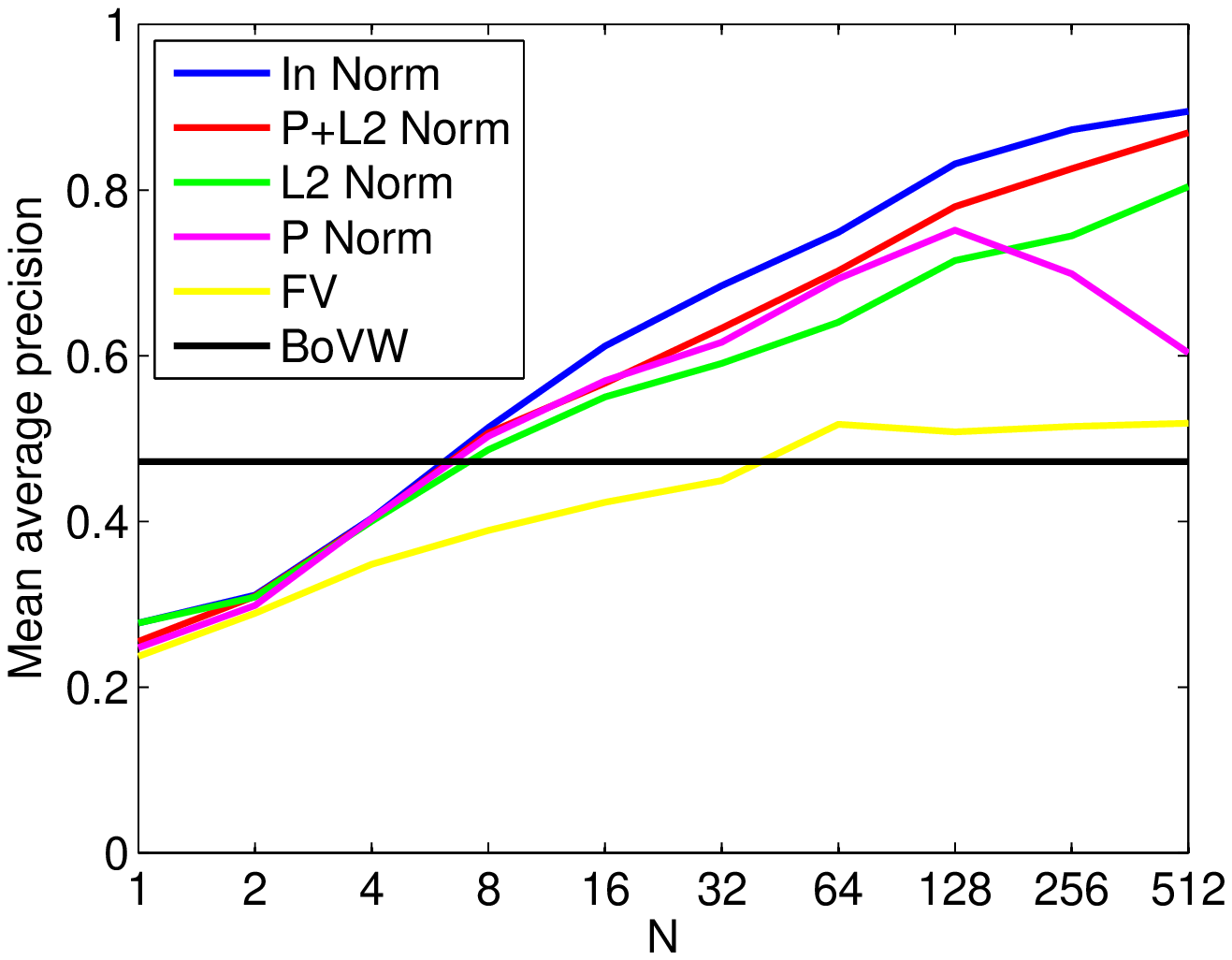} \\
		\centering (d) dvd
	\end{minipage}
	\begin{minipage}[b]{0.39\linewidth}
		\includegraphics[width=\linewidth]{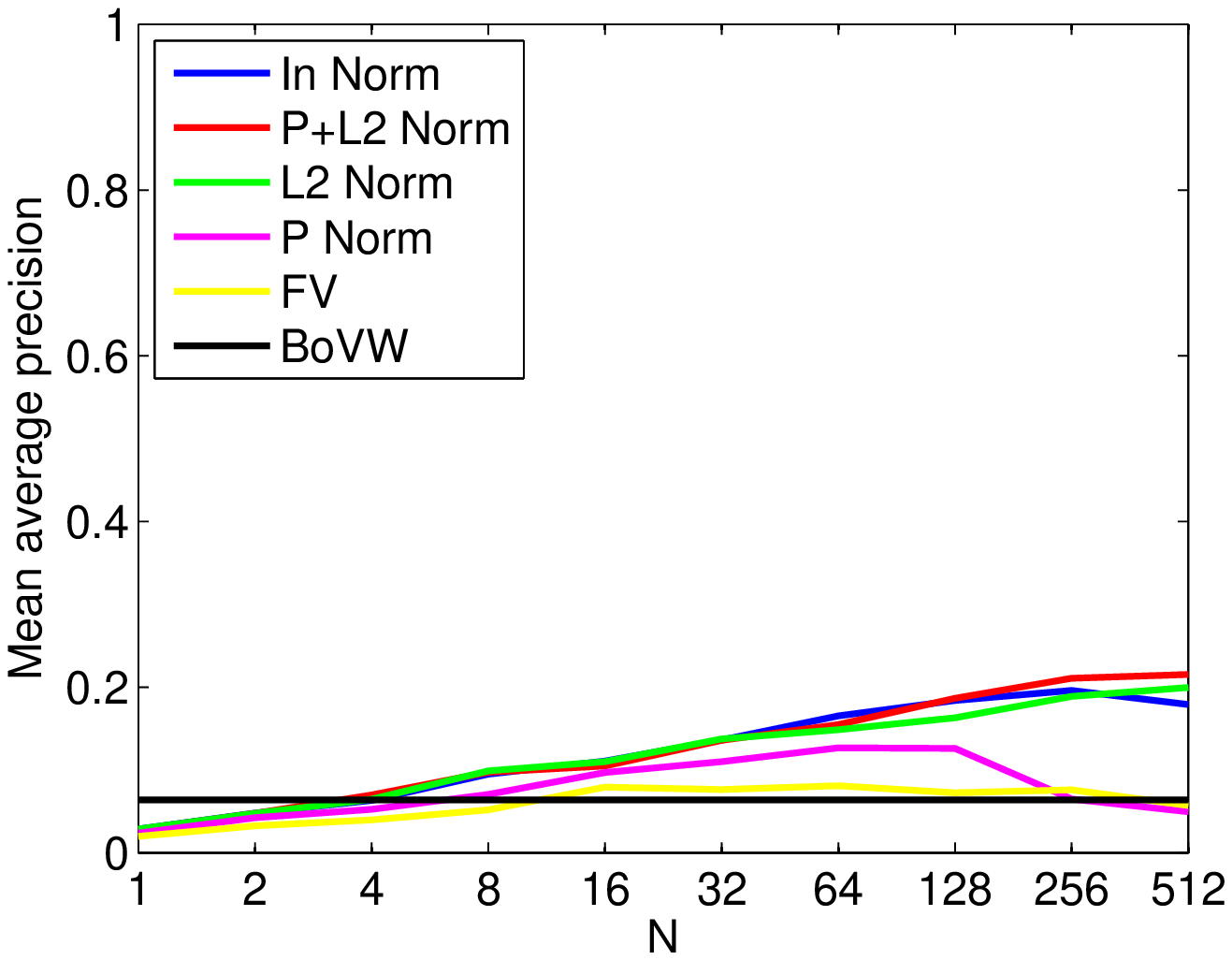} \\
		\centering (e) landmark
	\end{minipage}
	\begin{minipage}[b]{0.39\linewidth}
		\includegraphics[width=\linewidth]{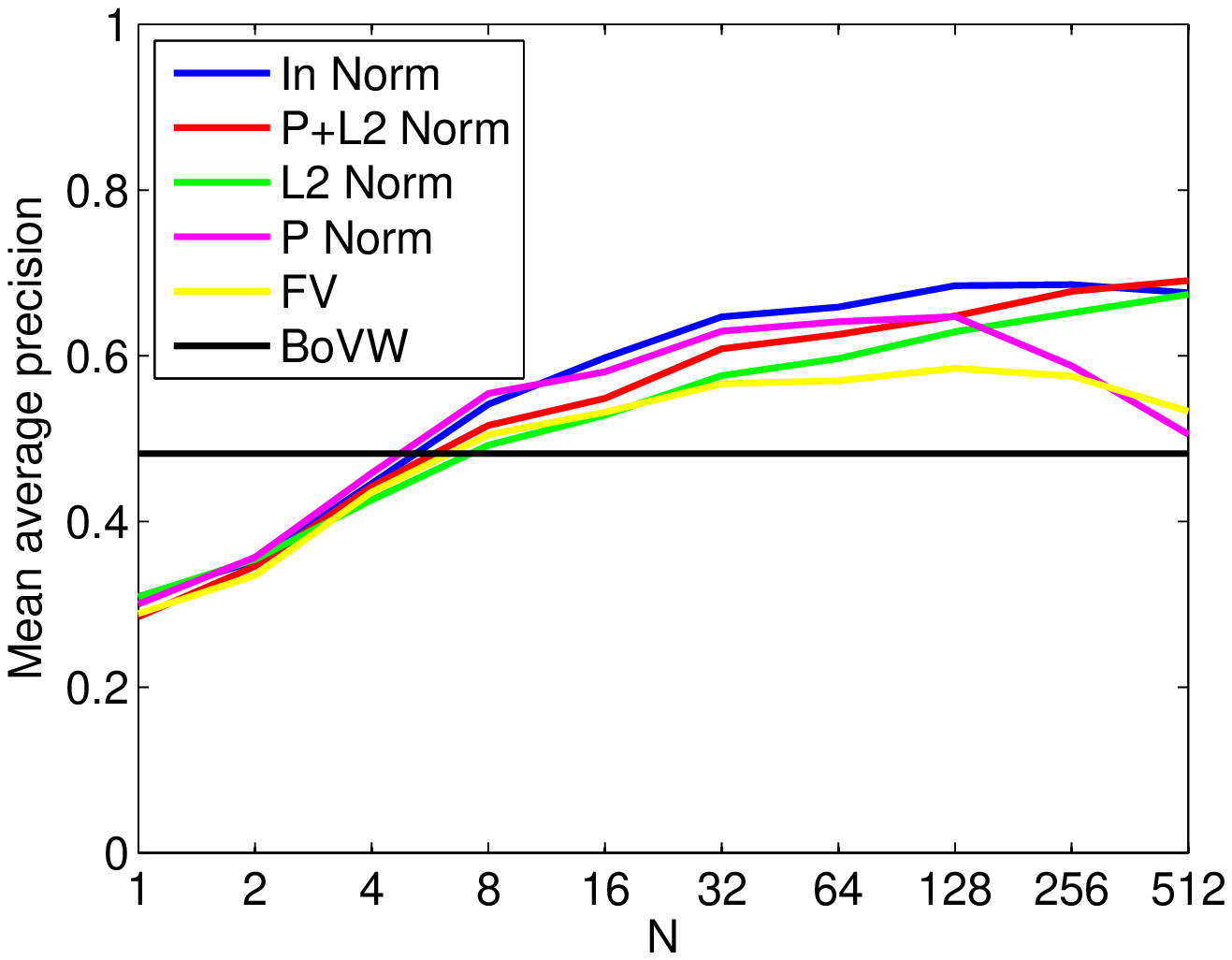} \\
		\centering (f) painting
	\end{minipage}
	\begin{minipage}[b]{0.39\linewidth}
		\includegraphics[width=\linewidth]{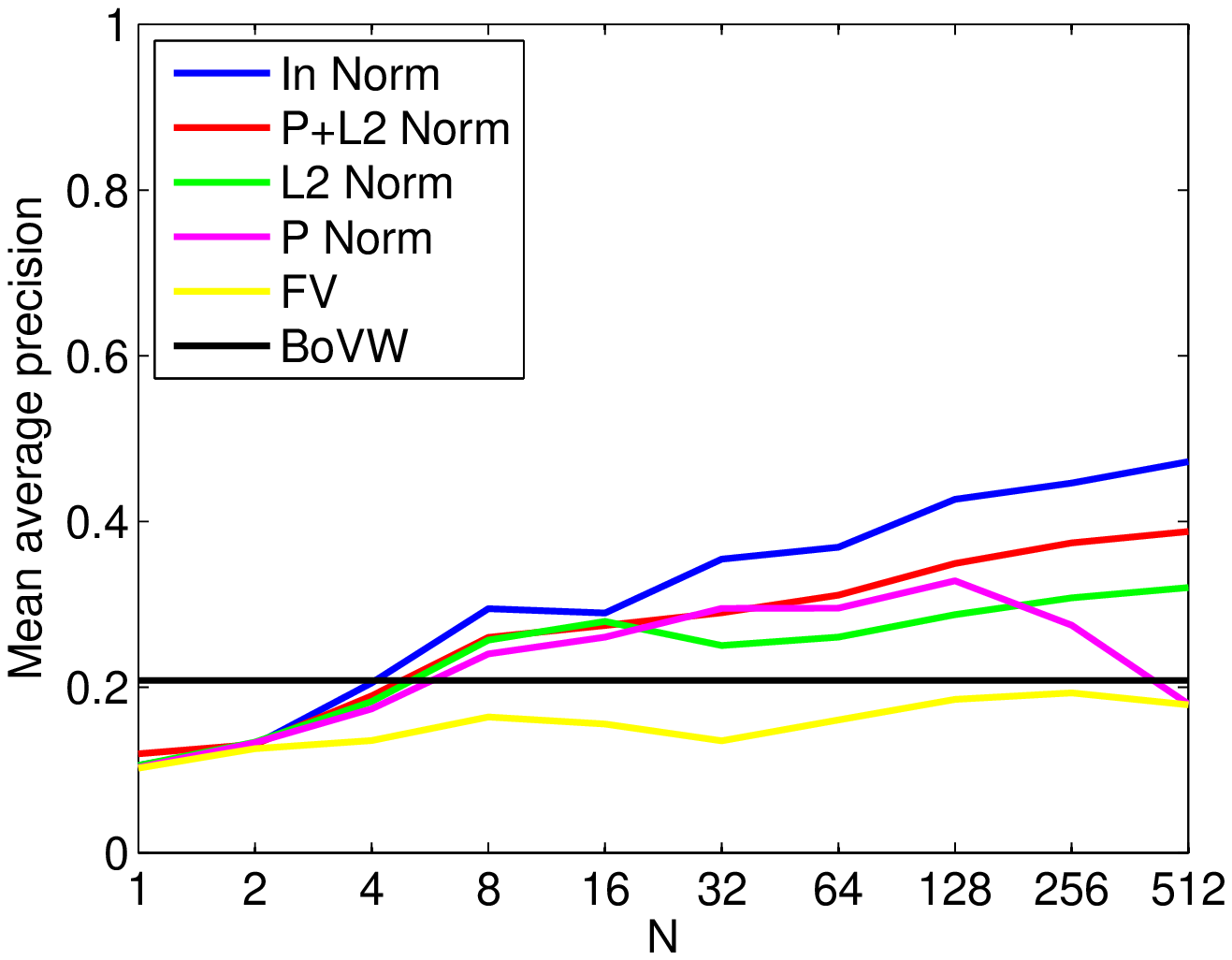} \\
		\centering (g) document
	\end{minipage}
	\begin{minipage}[b]{0.39\linewidth}
		\includegraphics[width=\linewidth]{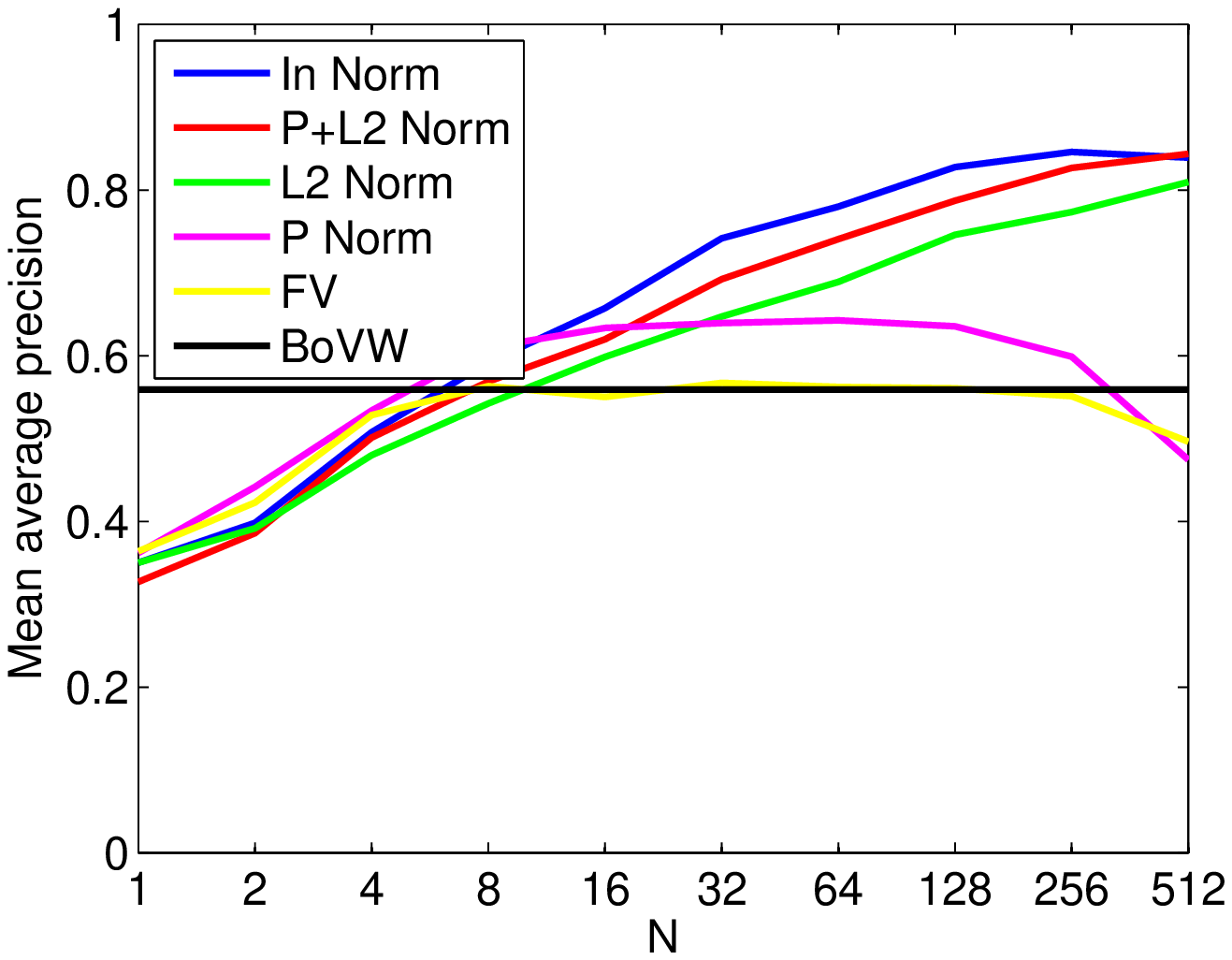} \\
		\centering (h) video
	\end{minipage} \\
	\caption{Comparison of the Fisher vector and BoBW representations applied to binary features on eight classes}
	\label{fig:map}
\end{figure*}

\subsection{Impact of Normalization}
The performance of the Fisher vector of binary features is evaluated in terms of image retrieval accuracy.
In particular,
the effect of the normalization methods described in Section~\ref{sec:normalization}
is investigated.
The following six methods are compared:
(1) bag of binary words approach with 1024 centroids (BoBW)~\cite{gal_iros11},
(2) Fisher vector without normalization (FV),
(3) Fisher vector with $\ell_2$ normalization (L2 Norm),
(4) Fisher vector with power normalization (P Norm), and
(5) Fisher vector with both power and $\ell_2$ normalization (P+L2 Norm).
(6) Fisher vector with intra normalization (In Norm).

Figure~\ref{fig:map} shows
a comparison of the Fisher vector and BoBW representations applied to binary features on eight classes.
The accuracy of the Fisher vector without any normalization (FV) is disappointing compared to the BoBW framework.
If $\ell_2$ or power normalization is adopted,
the accuracy of the Fisher vector is significantly improved.
The combination of the two normalizations further improves the performance,
which is consistent with the case of SIFT+GMM~\cite{per_eccv10}.
A little surprisingly, in many casese,
the intra normalization method outperforms the other normalization methods.
With appropriate normalization methods,
the accuracy improves as the number $N$ of components increases,
which is also consistent with the case of SIFT+GMM~\cite{jeg_pami12}.
Table~\ref{tab:sum_result} shows the accuracy of the proposed method (In Norm, $N = 512$) and the BoVW method on eight classes.
We can see that the proposed Fisher vector consistently outperforms the BoVW method on different datasets.
In particular, the difference of accuracy between the proposed method and the BoVW method is relatively larger for book, card, dvd, document, and video classes.
These classes includes many simple edges and corners (e.g., logo or text), and binary features extracted from these edges and corners are similar to each other.
In the case of BoVW method, these binary features tend to be quantized into the same VW and less discriminative.
On the other hand, the proposed Fisher vector can capture the "difference" from the components of BMMs; therefore it is more discriminative, resulting in better results.

\begin{table}[tb]
	\centering
	\caption{Comparison of the proposed method with the BoVW method on eight classes in terms of accuracy (MAP).}
	\label{tab:sum_result}
	\begin{tabular}{c|cccc} \hline
				&cd			&book		&card		&dvd	\\ \hline
		Prop	&0.785		&0.892		&0.377		&0.895	 \\
		BoVW	&0.623		&0.625		&0.179		&0.472	 \\ \hline \hline
				&landmark	&painting	&document	&video	\\ \hline
		Prop	&0.179		&0.676		&0.472		&0.840	 \\
		BoVW	&0.064		&0.482		&0.208		&0.559	 \\ \hline
	\end{tabular} \\
\end{table}

\subsection{Performance for Various String Lengths}
Next,
we investigate
the impact of the length of the binary strings on accuracy.
In this experiment,
the first $D'$ bits of the full 256-bit string are used.
This is reasonable
because, in the ORB algorithm,
binary tests are sorted according to their entropies;
the leading bits are more important.

\begin{figure}[tb]
	\centering
	\includegraphics[width=0.8\linewidth]{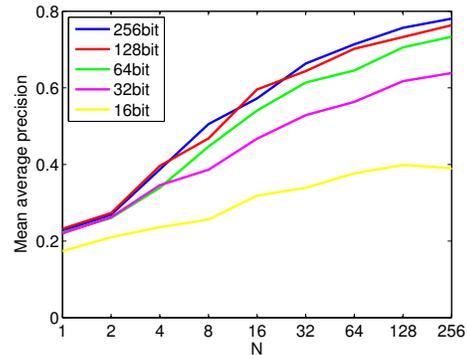} \\
	\caption{Accuracy of the Fisher vector representations under different string lengths.}
	\label{fig:bitnum}
\end{figure}

Figure~\ref{fig:bitnum} shows
the accuracy of the Fisher vector with intra normalization
as a function of the number of components $N$,
where the length of binary strings $D'$ varies from 16 to 256.
We can see that
the Fisher vector of longer binary strings achieves better accuracy.
However,
the gain becomes smaller
as the binary string becomes longer.
This is because the ending bits tend to be correlated to other bits
and thus are less informative.
Therefore, we can use shorter strings for efficiency at the cost of accuracy.
Because the computational cost of the Fisher vector is proportional to $N D'$\footnote{This is required in calculating the occupancy probability $\gamma_t (i)$, which is the most computationally expensive part of the propsoed Fisher vector as described in Section~\ref{sec:acceleration}.}, the other choice to reduce the computational cost is to use smaller $N$.
Figure~\ref{fig:bitnum} indicates that it is better to use shorter strings down to 64-bit strings instead of using smaller $N$.
For instance, the MAP score of 0.733 at $N = 256$ and $D' = 64$ is better than that of 0.714 at $N = 64$ and $D' = 256$.
Otherwise, it is better to use smaller $N$ instead of using smaller $D'$, e.g. 0.639 at $N = 256$ and $D' = 32$ v.s. 0.663 at $N = 32$ and $D' = 256$.
However, in this paper, we use full 256-bit strings in the other experiments to make the most of the ORB descriptor.

\subsection{Increasing Database Size}
\label{sec:distractor}
\begin{figure}[tb]
	\centering
	\includegraphics[width=0.8\linewidth]{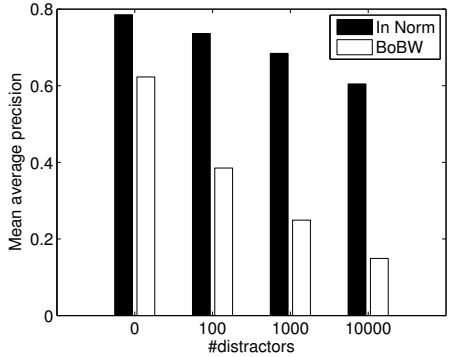} \\
	\caption{Comparison of the Fisher vector and BoBW representations with different numbers of distractors.}
	\label{fig:increase}
\end{figure}

We investigate the performance of the Fisher vector
when the size of the database becomes large.
In order to increase the size of database,
we use images in the MIR Flickr collection as a distractor in the same way as in~\cite{jeg_ijcv10}.
Figure~\ref{fig:increase}
compares the Fisher vector with intra normalization ($N = 512$, $D' = 256$) and the BoBW representation
with the different numbers of distractors.
In Figure~\ref{fig:increase},
0, 100, 1,000, and 10,000 distractor images are added to 100 reference images,
resulting that the size of database becomes 100, 200, 1,100, and 10,100 respectively.
We can see that
the Fisher vector achieves better performance for all database sizes.
Although the accuracy of both the Fisher vector and the BoBW representations
drops as the size of the database increases,
the degradation of the Fisher vector is relatively small.
This is because
the Fisher vector can represent higher order information than the BoBW representation
and is more discriminating even with larger database sizes.
It can be said that
the effectiveness of the proposed Fisher vector representation becomes more significant
when the database size is increased.

\subsection{Evaluation of Fast Approximated Fisher Vector}
\begin{figure}[tb]
	\centering
	\includegraphics[width=0.8\linewidth]{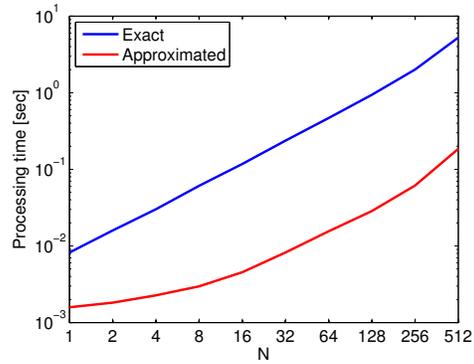} \\
	(a) Processing time as a function of $N$. \\
	\includegraphics[width=0.8\linewidth]{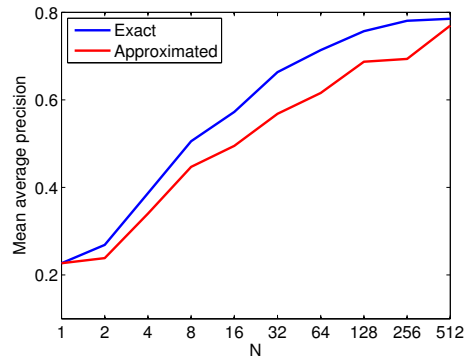} \\
	(b) Accuracy as a function of $N$. \\
	\caption{Evaluation of fast approximated Fisher vector.}
	\label{fig:time}
\end{figure}

Finally,
we evaluate
the performance of the approximated Fisher vector described in Section~\ref{sec:acceleration}.
Figure~\ref{fig:time}
compares the approximated and exact Fisher vector with intra normalization ($D' = 256$).
The distractor images in Section~\ref{sec:distractor} are not used in this experiment.
It can be seen that the approximated Fisher vector is
one order of magnitude faster than the exact Fisher vector
while the degradation of accuracy is only 6.4\% on average and 1.6\% for $N=512$.
Table~\ref{tab:approx_result} shows the average degradation of the MAP score on eight classes when the approximated Fisher vector with intra normalization ($D' = 256$) is adopted.
We can see that there is not much difference in the degradation of accuracy among different classes.

\begin{figure}[tb]
	\centering
	\includegraphics[width=0.8\linewidth]{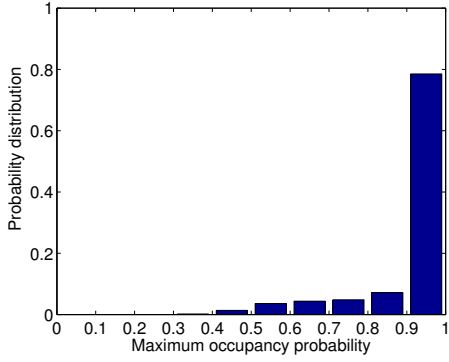} \\
	(a) The distribution of maximum occupancy probability $\max_i \gamma_t (i)$ for $N = 512$. \\
	\includegraphics[width=0.8\linewidth]{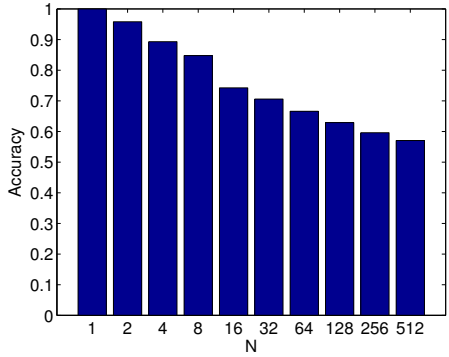} \\
	(b) The accuracy of the maximum occupancy probability approximation. \\
	\caption{The accuracy of two approximations used in the proposed Fisher vector.}
	\label{fig:peak}
\end{figure}

This approximated Fisher vector uses two approximations.
The first one is the approximation of the occupancy probability $\gamma_t (i)$;
we approximate $\gamma_t (i)$ with the largest value to 1 and the others to 0 as Eq.~(\ref{eq:approx2}).
This is based on the assumption that $\gamma_t (i)$ is peaky.
Figure~\ref{fig:peak} (a) shows the distribution of maximum occupancy probability $\max_i \gamma_t (i)$ for $N = 512$.
We can confirm that $\max_i \gamma_t (i)$ is near to 1 in most cases as expected.
The other approximation is that the component with maximum occupancy probability
is approximately obtained as $\hat{i} = \arg\min_i |x_t - y_i|$ using representative binary vectors defined in Eq.~(\ref{eq:approx1}).
Figure~\ref{fig:peak} (b) shows the accuracy of this approximation;
the probability that $\arg\min_i |x_t - y_i| = \arg\max_i \gamma_t (i)$.
We can see that although the accuracy declines as $N$ increases,
the accuracy of this approximation is still 57\% even for $N = 512$.
As the final approximated Fisher vector is created using a number of feature vectors,
this approximation works well as shown in Figure~\ref{fig:time}.

\begin{table}[tb]
	\centering
	\caption{Average degradation of the MAP score on eight classes when the approximated Fisher vector is adopted.}
	\label{tab:approx_result}
	\begin{tabular}{cccc} \hline
			cd			&book		&card		&dvd	\\ \hline
			6.4\%		&4.9\%		&6.8\%		&8.3\%	 \\ \hline \hline
			landmark	&painting	&document	&video	\\ \hline
			3.0\%		&4.1\%		&7.8\%		&3.9\% \\ \hline
	\end{tabular} \\
\end{table}

\section{Conclusions}
In this paper,
we proposed the application of the Fisher vector representation to binary features to improve the accuracy of binary feature based image retrieval.
We derived the closed-form approximation of the Fisher vectors of binary features that are modeled by the Bernoulli mixture model.
In addition, we also proposed a fast approximation method that accelerates the computation of the proposed Fisher vectors by one order of magnitude with comparable performance.
The effectiveness of the Fisher vectors of binary features was confirmed.
There were some interesting observations;
for example, the performance of the Fisher vector without power and $\ell_2$ normalization was very poor,
while the Fisher vector with power and $\ell_2$ normalization outperformed the BoBW framework.
The effectiveness of the proposed Fisher vector representation became more significant
when the database size increased.
Furthermore,
we demonstrated that
the intra normalization method originally proposed for VLAD also worked well for the proposed Fisher vector
and outperformed the conventional normalization methods.
This result encourages us to apply the intra normalization method to the Fisher vector of GMM.
In future,
we will apply the Fisher vector of binary features to
image classification problems.
We also expect that
the proposed Fisher vector representation can also be successfully applied
to other modalities such as audio signals.

\appendix
\label{sec:appendix}
We derive the Fisher information matrix under the following three assumptions:
(1) the Fisher information matrix $F_{\lambda}$ is diagonal,
(2) the number of binary features $x_t$ extracted from an image is constant and equal to $T$, and
(3) the occupancy probability $\gamma_t (i)$ is peaky.
From Eq.~(\ref{eq:fi}), we get:
\begin{eqnarray*}
F_{\mu_{id}}
&=&
\mathrm{E} \left[ \left(
\frac{\partial \mathcal{L} (X | \lambda)}{\partial \mu_{id}}
\right)^2 \right]
=
\mathrm{E} \left[ \left(
\sum_{t = 1}^T \frac{\partial \mathcal{L} (x_t | \lambda)}{\partial \mu_{id}}
\right)^2 \right] \\
&=&
\sum_{t = 1}^T
\mathrm{E} \left[ \left(
\frac{\partial \mathcal{L} (x_t | \lambda)}{\partial \mu_{id}}
\right)^2 \right] \\
&& +
2
\sum_{1 \le t < s \le T}
\mathrm{E} \left[
\frac{\partial \mathcal{L} (x_t | \lambda)}{\partial \mu_{id}}
\right]
\mathrm{E} \left[
\frac{\partial \mathcal{L} (x_s | \lambda)}{\partial \mu_{id}}
\right].
\end{eqnarray*}
If the parameter set $\lambda$ is estimated with maximum-likelihood estimation,
we have:
\begin{equation*}
\mathrm{E} \left[
\frac{\partial \mathcal{L} (x_t | \lambda)}{\partial \mu_{id}}
\right]
= 0.
\end{equation*}
Using the value of the Fisher score in Eq.~(\ref{eq:fs}), we get:
\begin{eqnarray*}
\label{eq:halfway}
\mathrm{E} \left[ \left(
\frac{\partial \mathcal{L} (x_t | \lambda)}{\partial \mu_{id}}
\right)^2 \right]
&=&
\int_{x_t} p(x_t | \lambda) 
\frac{\gamma_t^2 (i)}{\left( \mu_{id}^{x_{td}} (1 - \mu_{id})^{1 - x_{td}} \right)^2} d x_t \\
&=&
\int_{x_{td}=1} p(x_t | \lambda) 
\frac{\gamma_t^2 (i)}{\mu_{id}^2} d x_t \\
&&+
\int_{x_{td}=0} p(x_t | \lambda) 
\frac{\gamma_t^2 (i)}{(1 - \mu_{id})^2} d x_t.
\end{eqnarray*}
Using the assumption that the occupancy probability $\gamma_t (i)$ is peaky,
we approximate $\gamma_t^2 (i)$ as $\gamma_t (i)$.
Finally, using the following equations,
\begin{eqnarray*}
\int_{x_{td}=1} p(x_t | \lambda) \gamma_t (i) d x_t &=& w_i \sum_{j=1}^N w_j \mu_{jd}, \\
\int_{x_{td}=0} p(x_t | \lambda) \gamma_t (i) d x_t &=& w_i \sum_{j=1}^N w_j (1 - \mu_{jd}),
\end{eqnarray*}
we obtain:
\begin{equation*}
F_{\mu_{id}} =
T w_i
\left(
\frac{\sum_{j=1}^N w_j \mu_{jd}}{\mu_{id}^2}
+
\frac{\sum_{j=1}^N w_j (1 - \mu_{jd})}{(1 - \mu_{id})^2}
\right).
\end{equation*}

{\small
\bibliographystyle{ieee}
\bibliography{refs}
}

\end{document}